# Facial Expressions Recognition with Convolutional Neural Networks

**Subodh Lonkar**
learner.subodh@gmail.com

**ABSTRACT**

Over the centuries, humans have developed and acquired a number of ways to communicate. But hardly any of them can be as natural and instinctive as facial expressions. On the other hand, neural networks have taken the world by storm. And no surprises, that the area of Computer Vision and the problem of facial expressions recognitions hasn't remained untouched. Although a wide range of techniques have been applied, achieving extremely high accuracies and preparing highly robust FER systems still remains a challenge due to heterogeneous details in human faces. In this paper, we will be deep diving into implementing a system for recognition of facial expressions (FER) by leveraging neural networks, and more specifically, Convolutional Neural Networks (CNNs). We adopt the fundamental concepts of deep learning and computer vision with various architectures, fine-tune it's hyperparameters and experiment with various optimization methods and demonstrate a state-of-the-art single-network-accuracy of 70.10% on the FER2013 dataset without using any additional training data.

## I. INTRODUCTION

The idea or task of Facial Expressions Recognition (FER) can be thought of as: Given an input facial image, classify the underlying emotions or expressions into one of the basic emotions, like happiness, sadness, surprise, fear, anger, disgust or neutral. Because facial expressions are one of the most natural and instinctive indications of emotions, these categories of expressions are widely accepted in the area of facial expressions recognition. It finds a number of applications, some of the major ones include human-computer interactions, customer satisfaction, gaming industry and acquiring real-time human feedback. As it's a classification problem, techniques such as Support Vector Machines, Random Forest or Logistics Regression may be applied. In addition to these classical machine learning methods, neural networks have shown a lot of promise in solving this problem. Developments in computer vision have continuously improved upon the existing techniques and have shown great potential. Input data can be of multiple formats, like vectors, texts, sequences, images and various deep learning methods are required to handle different inputs and to solve the task under consideration. Although training systems for recognizing expressions under controlled conditions (eg. posed faces and frontal expressions) can be addressed with high accuracy, this problem still remains a challenge under dynamic lighting and pose conditions.

Convolutional Neural Networks (CNNs) perform well on image data. They have tremendous feature extraction capability and are computationally efficient too. They are the most widely deep learning models for image data. Images are just multiple matrices stacked onto one another. CNNs create a mechanism for extracting and analyzing minute details or features from images which informs a decision about each image as a whole. The Convolution operation in CNNs can be thought of as a simple extension to dot product on vectors, mathematically, it's multiplication followed by addition. The dot product is applied on vectors while CNNs are applied on matrices. As weights are learnt in an MLP, kernels or kernel matrices are learnt in CNN. Apart from the fundamental concepts of neural networks like activation functions, backpropagation, dropouts, optimizers and initializers, CNNS also leverages some other ideas like Kernels, padding, strides, pooling and data augmentation. The idea and working behind each of these concepts are out of the scope of this research paper and can be read from other resources. In this work, we aim to improve the prediction accuracy by constructing various experiments and explore different ways to fine-tune our model and training hyperparameters for best results and finally achieve a state-of-the-art single-network-accuracy of 70.10% on the FER2013 dataset without using any additional training data.

Finally, we deploy our model by building a simple web application which will give us the recognition results on taking images as inputs.

## II. DATASETS

**The Extended Cohn-Kanade Dataset (CK+)**

This dataset was primarily released for the purpose of promoting research into automatically detecting individual facial expressions. It contains 593 mostly gray image sequences (327 sequences having discrete emotion labels) of 123 subjects. The ground truth consists of facial expression labels and FACS (AU label for final frame in each image sequence). The dataset contains images with a resolution of 640x490 with posed and spontaneous smiles. The underlying facial expressions belong to the following categories: neutral, sadness, surprise, happiness, fear, anger, contempt and disgust.

**Japanese Female Facial Expressions (JAFFE)**

It contains 213 static grayscale images of 10 subjects with a resolution of 296x296. Images are posed with facial expressions like neutral, sadness, surprise, happiness, fear, anger, and disgust as the ground truth.

**FERG-3D-DB (Facial Expression Research Group 3D Database) for stylized characters**

This dataset contains 39574 color annotated examples of 4 subjects with emotion labels. Images are mostly in frontal pose and contain seven expressions: angry, disgust, fear, joy, neutral, sad, surprise.

**Radboud Faces Database (RaFD)**

This is a unique dataset that contains color images captured from three different gaze directions and five camera angles. A total of 8040 posed images of 67 subjects with a resolution of 681x1024 with emotion labels are present.

**FER2013 Dataset**

The dataset contains 35887 images normalized to 48x48 pixels in grayscale. It is a well-studied dataset and is widely used in the field of recognizing facial expressions. It contains seven categories of expressions, namely, happy(8,989), sad(6,077), surprise(4,002), angry(4,953), disgust(547), fear(5,121) and neutral(6,198). As can be seen, it contains an imbalance in the number of images belonging to each class and is one of the most challenging datasets to work on. Images are marked as Train, private test and public tests as per its usage.

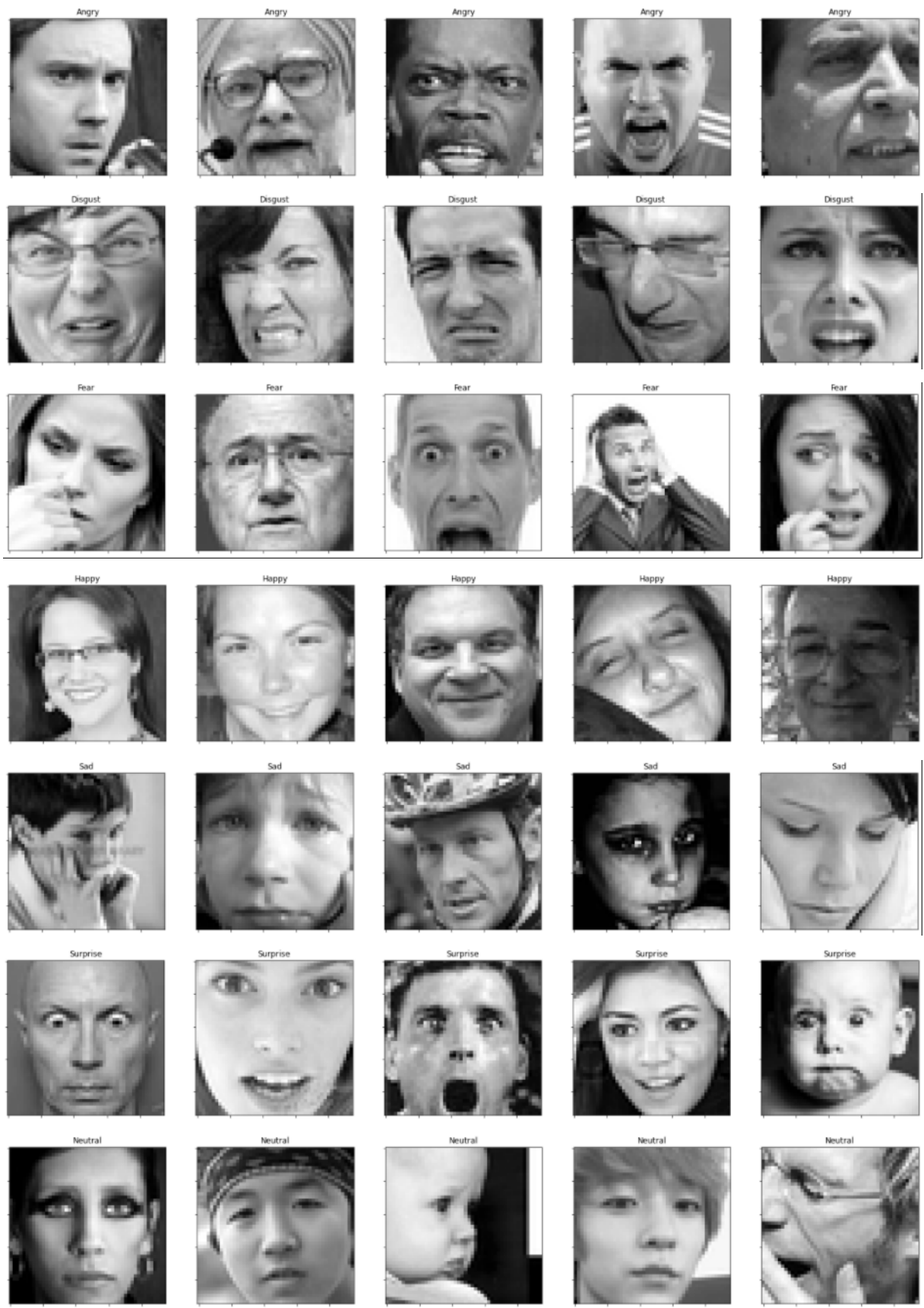

Figure 1: Sample images from the FER2013 Dataset

# III. METHOD

The task here is to classify an input image with some expression, we have to classify it into one of the seven emotions or expressions. Thus, classification techniques like Support Vector Machines, Logistic Regression or the Random Forest can be applied. But we will be using a deep learning technique here, which makes use of CNNs.

I evaluated a bunch of preprocessing techniques followed by tuning the architecture and finally achieved a nearly state-of-the-art accuracy of 70.10% on the test data. Let's have a brief look at some of the techniques applied. Refer Figure 3 for the custom model architecture.

**Data Augmentation**

It is a technique used to increase the diversity of the training data by applying random (but realistic) transformations such as image rotation, sheer, flip, rotation, scale, addition of noise, crop, zoom and brightness. Here, considering the use case at hand, I applied augmentation techniques like horizontal flip, shear, rotation, scaling, zoom in and zoom out as the face and the underlying expressions can be at different distances and changes in brightness level as the input image can be captured in different illumination conditions. This showed a considerable improvement in the prediction accuracy as it helped the model to generalize better for slight changes in input data. But remember, while data augmentation does a lot of good, it has its downsides too. It can unknowingly enhance the already existing bias in input data which might hamper the model performance.

**Optimizer**

I experimented with some popular optimizers like SGD, SGD with momentum, Adam, NAdam and Adamax. SGD and SGD with momentum showed good performance but were observed to be relatively slow for converging. Adam and NAdam performed the best, the difference in results wasn't significant, I went with NAdam as it is a more popular approach.

**Activation Function**

As it's a seven class classification problem, I used a softmax in the last layer which outputs one of the most probable classes from the seven classes. For hidden layers, I used the relu activation as it avoids the problem of vanishing and exploding gradients. As he initializers are observed to work well with relu activations, I used the he normal initializer in this case.

**Dropout**

Dropouts act as a regularization technique. It reduces the risk and dependency on single neurons firing for certain emotions. It brings in regularization by randomization. As it is very easy to overfit with neural networks, dropouts with appropriate dropout rates were applied. It acts as a weak regularizer and improves the generalization capabilities of the model.

**Batch Normalization**

Training deep neural networks with a large number of layers is challenging as they can be sensitive to the initial random weights and configuration of the learning algorithm. One possible reason for this difficulty is the distribution of the inputs to layers deep in the network may change after each mini-batch when the weights are updated. This might cause the learning algorithm to forever chase a moving target. This change in the distribution of inputs to layers in the network is referred to the technical name "internal covariate shift". Batch normalization is a technique for training very deep neural networks that standardizes the inputs to a layer for each mini-batch. This has the effect of stabilizing the learning process and dramatically reducing the number of training epochs required to train deep networks. Batch normalization provides an elegant way of reparametrization of almost any deep network. The reparametrization significantly reduces the problem of coordinating updates across many layers. It primarily reparametrize the model to make some units always be standardized by definition. Batch Normalization has been used in middle layers to address this issue.

**Class Weighting**

To overcome the problem of class imbalance, I used class weights inversely proportional to the number of samples present in the given class. Refer Figure 2 for better look at the class imbalance.

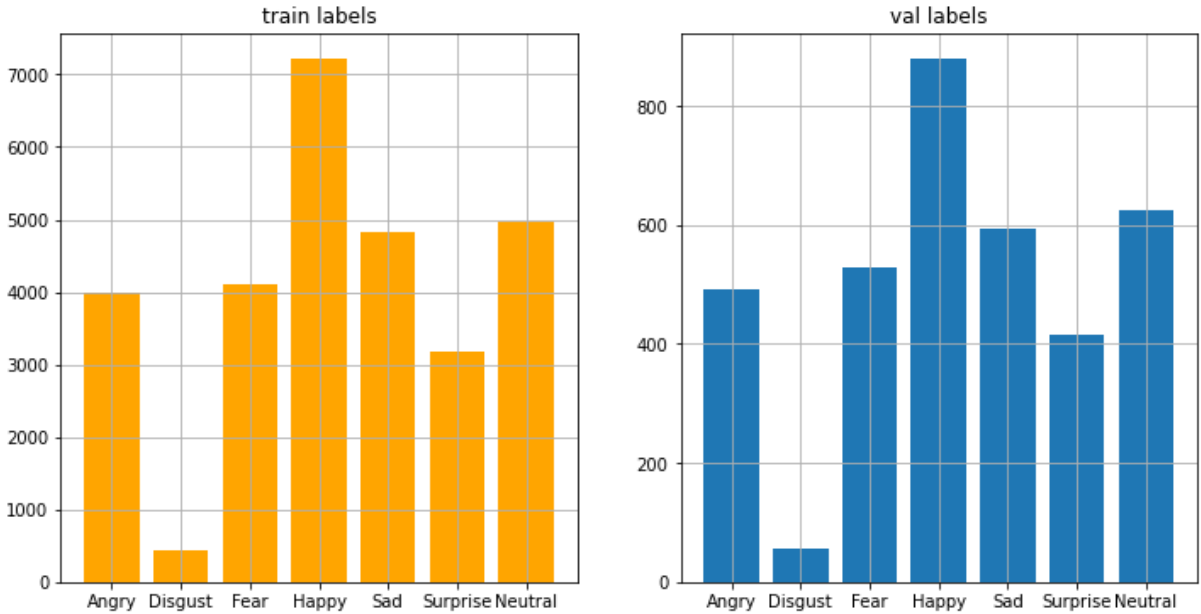

Figure 2: Class imbalance in train (Training) and validation (Public Test) data

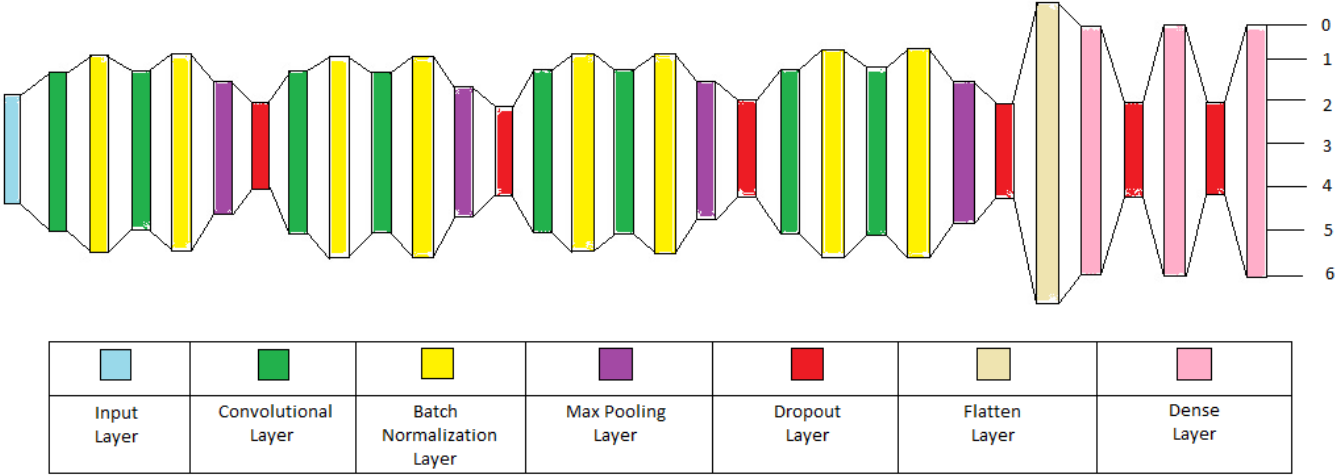

Figure 3: CNN Custom Architecture

## IV. RESULTS

The evaluation metric that we are using here is Accuracy. The final single-network-model achieved an accuracy of 70.10% on the test data.

For error analysis, I target confusion matrix scores. Around 90% of the images with happy labels were classified as happy while the prediction accuracy for the class Fear was just under 50%. Class Happy had the highest precision and recall while the class Fear had the least precision and recall among all the seven classes. The model performed extremely well on classes Happy, Surprise and Neutral while it got a bit confused with emotions such as fear and sadness where it didn't return that good scores.

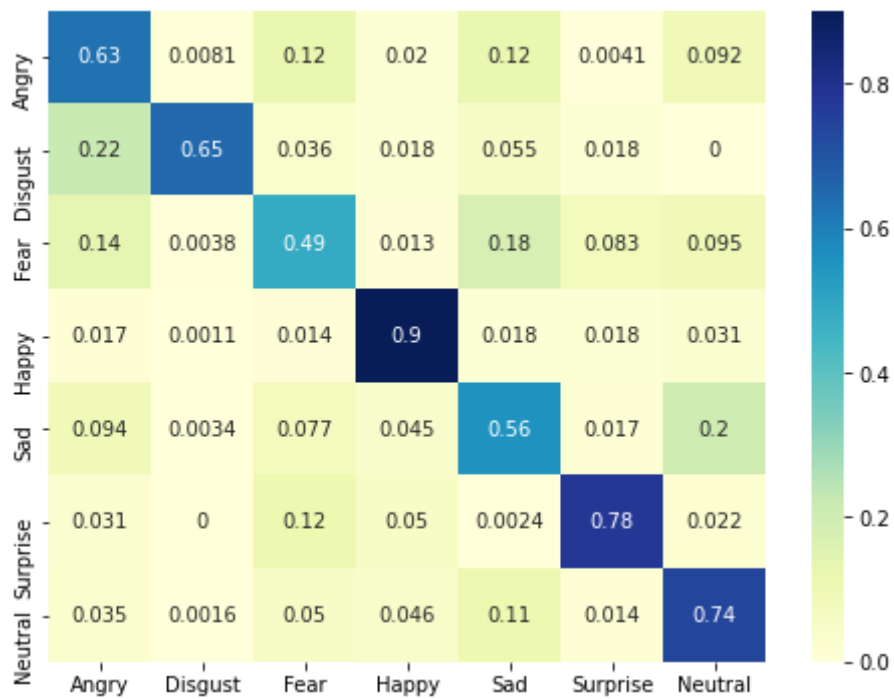

Figure 4: Confusion Matrix

If we look at the accuracy and loss plots, we can observe that the model started to overfit after the 40th epoch as the gap between the train and test curves started widening as the model started capturing some irrelevant details. I used callbacks like EarlyStopping to avoid the model from learning too many insignificant details and ModelCheckpoint for saving the best models in terms of loss and accuracy.

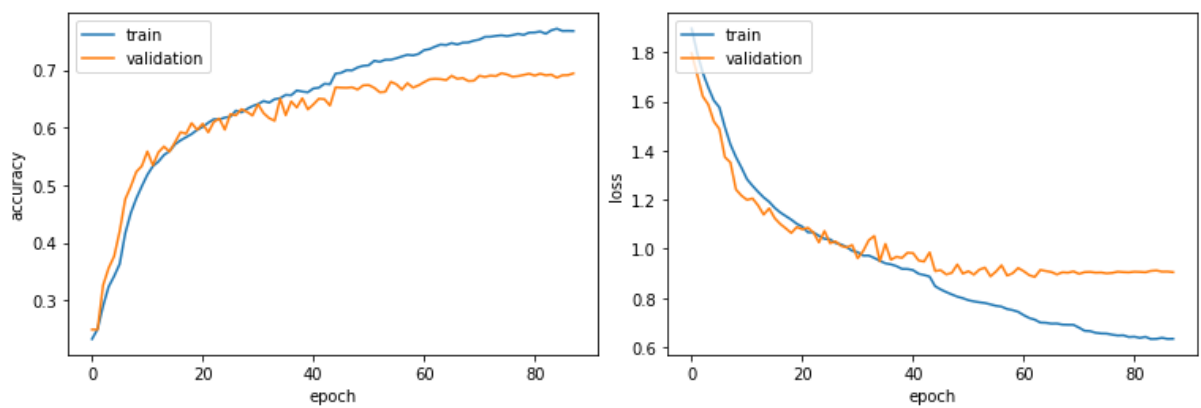

Figure 5: Accuracy vs Epoch and Loss vs Epoch plots

## V. CHALLENGES

It can be a bit difficult to perform the error analysis since the trained model performed better than human-level accuracy and analyzing errors was quite difficult for classes such as Fear. Because emotions are subjective in nature, it is often the case that images might have multiple interpretations. Thus, it might be sometimes difficult to differentiate between emotions such as fear and sadness or happiness and surprise.

## VI. WEB APPLICATION

I also built and deployed a web application which uses the above trained model in the backend to make predictions and performs prediction in real time with negligible latency.

Link to demonstration of working of the web app: https://youtu.be/Tx-iHP9KY5w

## VII. CONCLUSION

In this research paper, I designed and trained my own custom CNN architecture on the FER2013 dataset that achieved state-of-the-art single-network-accuracy of 70.10% on the test data. It included applying the techniques of image augmentation followed by fine tuning the model architecture and it's hyperparameters. We leveraged the fundamental ideas of neural networks and specifically CNNs, like dropouts, batch normalization, padding, pooling, strides, kernel size and number of kernels to use, activation function, weight initialization techniques, various optimizers, padding and pooling. The proposed model architecture outperforms many state-of-the-art methods and remains very much competitive. I have used the data as per the given usage, that is, Training, Public test and Private Test for training, validation and testing respectively. Randomly splitting these datasets can result in much better results, but here I have stuck to using them as per the mentioned usage.. I overcame the issue of class imbalance by using class weights. Additionally, I demonstrated that Facial Expressions Recognition models can be applied in real world scenarios by developing a web application.

## VIII. FUTURE WORK

I have designed and trained my own cutomer CNN architecture. The designed model can also be better tuned by trying out some more combinations of potential parameters. As there is a lot of good research undertaken in this area, leveraging transfer learning can also be a good option to see if we can achieve any significant improvement over the existing solution. Other techniques like the facial landmark detection and alignment or attentional CNNs can also be implemented to achieve better accuracies on the test data.

## IX. REFERENCES

[1] Minh-An Quinn, Grant Sivesind and Guilherme Reis: "Real-time Emotion Recognition From Facial Expressions", CS 229 - Stanford University

[2] Darshan Gera, S Balasubramanian: Affect Expression Behaviour Analysis In The Wild Using Consensual Collaborative Training, arXiv preprint arXiv:2107.05736v1, 2021.

[3] James A Russell. A circumplex model of affect. Journal of personality and social psychology, 39(6):1161, 1980.

[4] Nitish Srivastava, Geoffrey Hinton, Alex Krizhevsky, Ilya Sutskever and Ruslan Salakhutdinov: "Dropout: A Simple Way to Prevent Neural Networks from
Overfitting", Journal of Machine Learning Research 15 (2014) 1929-1958.

[5] Amil Khanzada, Charles Bai and Ferhat Turker Celepcikay: "Facial Expression Recognition with Deep Learning", Stanford University - CS230 Deep Learning

[6] Anuvabh Dutt, Denis Pellerin, and Georges Quénot. Coupled ensembles of neural networks. Neurocomputing, 396:346–357, 2020

[7] Paul Ekman. Facial action coding system (facs). A human face, 2002.

[8] S. Li and W. Deng, "Deep facial expression recognition: A survey", arXiv preprint arXiv:1804.08348, 2018.

[9] Y. Tang, "Deep Learning using Support Vector Machines," in International Conference on Machine Learning (ICML) Workshops, 2013.

[10] B.-K. Kim, S.-Y. Dong, J. Roh, G. Kim, and S.-Y. Lee, "Fusing Aligned and Non-Aligned Face Information for Automatic Affect Recognition in the Wild: A Deep Learning Approach," in IEEE Conf. Computer Vision and Pattern Recognition (CVPR) Workshops, 2016, pp. 48–57.

[11] Quinn M., Sivesind G., and Reis G., "Real-time Emotion Recognition From Facial Expressions", 2017

[12] K. Simonyan and A. Zisserman, "Very deep convolutional networks for large-scale image recognition," in 3rd International Conference on Learning Representations, ICLR 2015 - Conference Track Proceedings, 2015.

[13] R. T. Ionescu, M. Popescu, and C. Grozea, "Local Learning to Improve Bag of Visual Words Model for Facial Expression Recognition," Work. challenges Represent. Learn. ICML, 2013.

[14] S. Minaee and A. Abdolrashidi, "Deep-emotion: facial expression recognition using attentional convolutional network," arXiv. 2019, doi: 10.3390/s21093046.

[15] A. A. Lydia and F. S. Francis, "Adagrad-An Optimizer for Stochastic Gradient Descent", 2019.

[16] B. Fasel and J. Luettin, "Automatic facial expression analysis: A survey," Pattern Recognition, vol. 36, no. 1. 2003, doi: 10.1016/S0031-3203(02)00052-3.

[17] I. J. Goodfellow, D. Erhan, P. L. Carrier, A. Courville, M. Mirza, B. Hamner, W. Cukierski, Y. Tang, D. Thaler, D.-H. Lee et al., "Challenges in representation learning: A report on three machine learning contests," in the International Conference on Neural Information Processing. Springer, 2013, pp. 117–124.

[18] A. Mollahosseini, D. Chan, and M. H. Mahoor, "Going Deeper in Facial Expression Recognition using Deep Neural Networks," CoRR, vol. 1511, 2015.

[19] Zhanpeng Zhang, Ping Luo, Chen Change Loy, and Xiaoou Tang, "From facial expression recognition to interpersonal relation prediction," International Journal of Computer Vision, Springer, vol. 126, no. 5, pp. 550–569, 2018.

[20] Xiaofeng Liu, BVK Vijaya Kumar, Jane You, and Ping Jia, "Adaptive deep metric learning for identity aware facial expression recognition," in Proceedings of the IEEE Conference on Computer Vision and Pattern Recognition Workshops, 2017, pp. 20–29.

[21] Vikas Gupta. Face detection – opencv, dlib and deep learning ( c++ / python ), Oct 2018.

22. François Chollet. Xception: Deep learning with depth wise separable convolutions. arXiv:1610.02357, 2016.

[23] Christopher Pramerdorfer and Martin Kampel. Facial expression recognition using convolutional neural networks: State of the art. 12 2016.

[24] S.S. Kilkarni, N.P. Reddy, and S. Hariharan. Facial expression (mood) recognition from facial images using committee neural networks. BioMed Eng OnLine, (16), 08 2009.